\documentclass[conference]{IEEEtran}
\IEEEoverridecommandlockouts
\usepackage{cite}
\usepackage{amsmath,amssymb,amsfonts}
\usepackage{algorithmic}
\usepackage{graphicx}
\usepackage{textcomp}
\usepackage{xcolor}
\usepackage{url}
\usepackage{multirow}
\usepackage[english,vietnamese]{babel}

\usepackage{tikz}
\usetikzlibrary{arrows.meta}

\definecolor{myblue}{RGB}{55, 222, 222}
\definecolor{mygreen}{RGB}{80,160,80}
\definecolor{myred}{RGB}{160,80,80}
\definecolor{myyellow}{RGB}{160,160,80}
\definecolor{mycyan}{RGB}{232, 181, 60}
\definecolor{mypurple}{RGB}{160,80,160}
\definecolor{mygray}{RGB}{128,128,128}
\definecolor{myorange}{RGB}{255,128,0}
\definecolor{mybrown}{RGB}{128,64,0}

\def\BibTeX{{\rm B\kern-.05em{\sc i\kern-.025em b}\kern-.08em
    T\kern-.1667em\lower.7ex\hbox{E}\kern-.125emX}}
\usepackage[numbers]{natbib}
\begin{document}

\title{NOWJ1@ALQAC 2023: Enhancing Legal Task Performance with Classic Statistical Models and Pre-trained Language Models\\
}


\author{
Tan-Minh Nguyen$^1$, Xuan-Hoa Nguyen$^1$, Ngoc-Duy Mai$^1$, Minh-Quan Hoang$^1$, \\
Van-Huan Nguyen$^1$, Hoang-Viet Nguyen$^1$, Ha-Thanh Nguyen$^2$, Thi-Hai-Yen Vuong$^{*,1}$ \\ 
\textit{$^1$VNU University of Engineering and Technology}, Hanoi, Vietnam \\ 
\textit{$^2$National Institute of Informatics}, Tokyo, Japan \\ 
$^1$\{20020081, 21020072, 21020512, 21020553, 21021501, 20020160, yenvth\}@vnu.edu.vn \\
$^2$nguyenhathanh@nii.ac.jp
}

\maketitle
\selectlanguage{english}
\begin{abstract}
This paper describes the NOWJ1 Team's approach for the Automated Legal Question Answering Competition (ALQAC) 2023, which focuses on enhancing legal task performance by integrating classical statistical models and Pre-trained Language Models (PLMs). For the document retrieval task, we implement a pre-processing step to overcome input limitations and apply learning-to-rank methods to consolidate features from various models. The question-answering task is split into two sub-tasks: sentence classification and answer extraction. We incorporate state-of-the-art models to develop distinct systems for each sub-task, utilizing both classic statistical models and pre-trained Language Models. Experimental results demonstrate the promising potential of our proposed methodology in the competition.
\end{abstract}

\begin{IEEEkeywords}
natural language processing, information retrieval, question answering, machine learning, deep learning.  
\end{IEEEkeywords}

\section{Introduction}
Along with the increasing number of legal documents and the need for legal services, lawyers and law practitioners may need automated tools to support them in operations, including legal searching, contract review, and question answering. This can save time and human resources on paperwork while reducing the risk of human error.

The Automated Legal Question Answering Competition (ALQAC) \cite{thanh2021summary,nguyen2022alqac} is an annual competition to support the research community in the legal AI domain. This year, the competition included two subtasks: legal document retrieval and legal question answering. Legal retrieval is the process of searching relevant articles in response to a given question. Legal question answering aims to extract information from one or more legal documents. 

This paper describes our approaches to these tasks in the competition. For the document retrieval task, we applied a pre-processing step to overcome the input limitation. Afterward, learning-to-rank methods are employed to merge features from different models. For the question-answering task, we considered the task as two smaller problems: sentence classification and answer extraction. We utilized state-of-the-art models to construct two different systems for these problems. Experimental results show the promise of our proposal in the competition. 

The remainder of the paper is structured as follows: In Section 2, we present related works. A description of the methods of the two tasks is provided in Section 3. Section 4 shows the detailed experiments and results. Finally, we conclude the paper in Section 5 by summarizing our proposal and describing future work.

\section{Related Works} 
\subsection{Legal Document Retrieval}
In recent times, there has been remarkable progress in Natural Language Processing (NLP), exemplified by revolutionary models like BERT and GPT, which have garnered significant public interest. These cutting-edge NLP models have opened up new avenues for addressing the complex challenge of Legal Document Retrieval. Researchers have harnessed the power of these advanced models to devise efficient techniques for handling legal texts. These approaches prioritize not only accuracy but also the model's practical applicability within real systems. 

In Information Retrieval, Neural Models can be broadly categorized into two types: representation-focused and interaction-focused \cite{guo2020deep}. Representation-focused approaches learn separate representations for queries and documents and compare them using functions like cosine similarity or dot product. While this reduces retrieval time, representing the entire text with a single vector may not capture all crucial information. On the other hand, interaction-focused approaches compute word-by-word similarity matrices to generate relevance scores. Well-known methods like DRMM and convKNRM perform well in traditional ad-hoc retrieval tasks by effectively comparing queries and documents. Moreover, some neural models combine both representation-focused and interaction-focused approaches.

Since the success of BERT \cite{devlin2018bert}, the pre-trained language models have drawn great attention in the field of legal document retrieval. The BERT-PLI model \cite{shao2020bert}, which incorporates paragraph-level interaction into the BERT framework, is tailored for legal documents. However, a noteworthy limitation is its struggle with long-distance attention, leading to a performance impact, primarily caused by the text being broken down into paragraphs.

At COLIEE 2021, Amin Abol \cite{amin2018bert} made significant improvements to the retrieval effectiveness of the BERT re-ranker by combining document-level representation learning objectives with the ranking objective during the fine-tuning process (known as Multi-task optimization). As a result of these enhancements, the model achieved superior retrieval performance compared to the original BERT re-ranker while using the same training instances and structure

\subsection{Legal Question Answering}
Automated Question Answering in the legal domain has long since been regarded as one of the most impactful both to academic and business models. Yet only recently does it receive the attention it well deserves. In the early days, most approaches proposed are NLP-based techniques in order to find out the relevant information that can be used as an answer \cite{Kim2014}. However, with the recent advancement in neural networks, more and more researchers are turning to neural-based approaches, notably the family of Bert-based models \cite{Zhang_2021,holzenberger2020}. These approaches are still far from perfect. While traditional NLP techniques require a properly tuned workflow pipeline and face many difficulties in adapting to new data variance. Deep neural models have poor interpretability and require a huge amount of resources for training and generalizing to new data.

The Multiple Choice task involves providing the correct answer among a set of options for each question. In study \cite{chen2023legal}, the authors tackled the challenge of incorporating additional information from reference books, enabling the model to select the correct answer through sentence embedding technique and attention mechanism. To address the issue of data imbalance, Radha Chitta et al. proposed generating synthetic instances from the minority answer using the SMOTE algorithm \cite{chitta2019reliable}. Experimental results demonstrated the effectiveness of this method with imbalanced datasets.

\section{Method}
\subsection{Legal Document Retrieval}

Given a question $q$, our objective for this task is to build a system that is able to look up in a legal corpus $D={d_1,d_2,...,d_n}$, which contains a huge number of legal articles issued by the authorities, and provide articles that can answer to the question $q$. The accuracy of the built system can be evaluated by Formula \ref{eq:retrieval}:
\begin{equation}
    d^*=\underset{d \in D}{\operatorname{argmax}} R(q, d)
    \label{eq:retrieval}
\end{equation}
where $R(q, d)$ denotes the relevance score between the query $q$ and article $d$.

\begin{figure}[ht]
\centering
\includegraphics[width=0.8\linewidth]{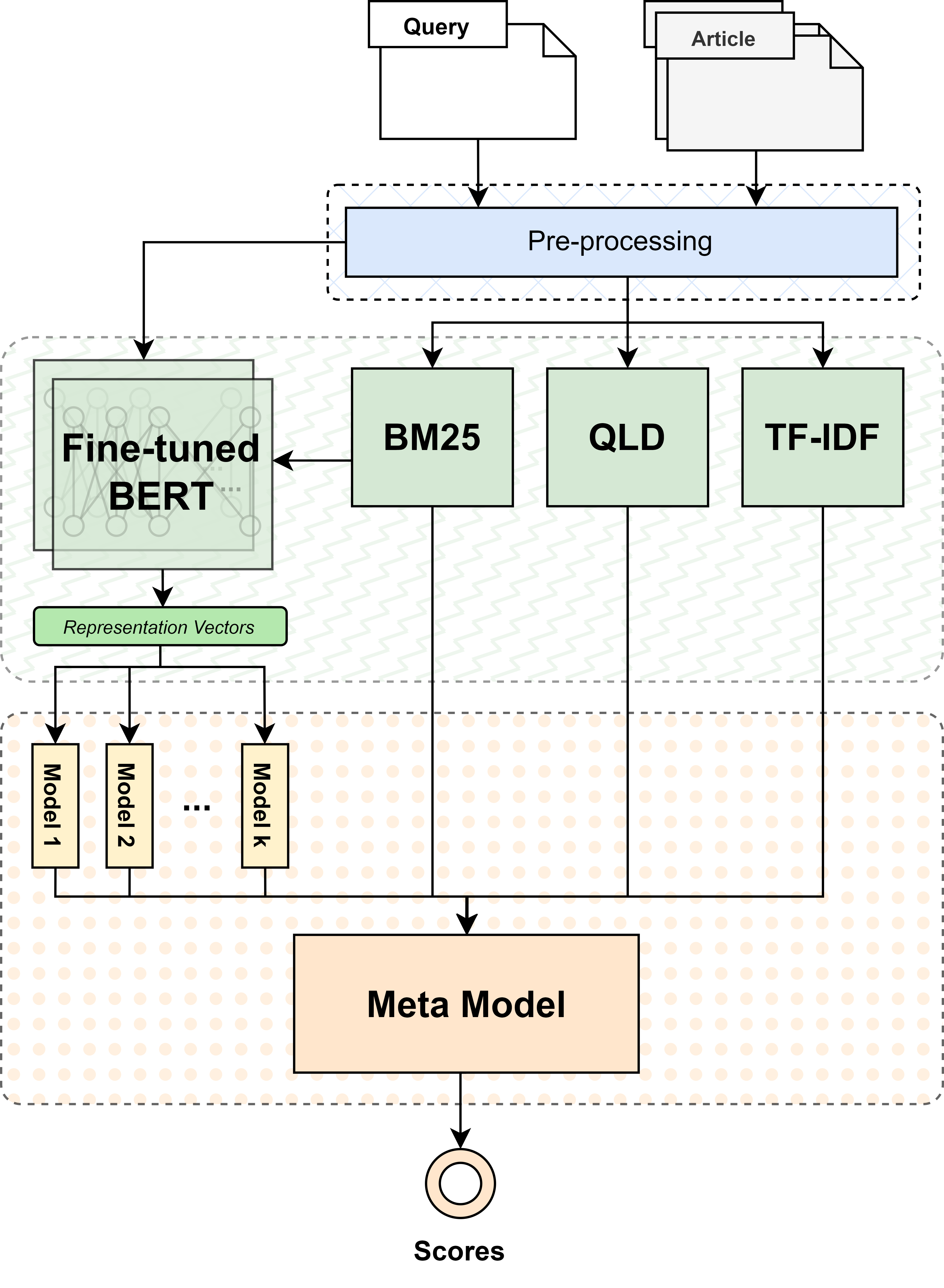}
\caption{Overview of our architecture for Task 1} 
\label{fig:task1_overview}
\end{figure}

For task 1, we propose an architecture combining traditional statistical models and BERT-based models. Specifically, we first perform a pre-processing of the data. We then implement traditional statistical models and BERT-based models. Multiple features are extracted for each pair of questions and candidates. Learning-to-rank methods are employed to aggregate these features into the final scores. Figure \ref{fig:task1_overview} illustrates details of our architecture. 

\subsubsection{Pre-processing} 
\selectlanguage{vietnamese}
Prior to the training process, we undertake the following pre-processing steps: 
\begin{itemize}
    \item Remove useless characters and abbreviations: We remove redundant characters such as newline characters and extra spaces. We then correct abbreviations such as convert \textit{BLDS} into \textit{Bộ luật dân sự}.
    \item Word segmentation: A Vietnamese word can be a compound word that is made of more than one single word. In some cases, two nonsense words can make up a meaningful compound word. Thus, word segmentation is performed to form compound words using RDRSegmenter \cite{NguyenNVDJ2018} to enhance the performance of traditional statistical models.
    \item Article segmentation: With the length, complication of legal text and model input limitation, it would be ineffective to feed raw article legal into these matching models. Thus, articles are segmented into smaller units, such as titles, clauses and sub-clauses.
\end{itemize}

\selectlanguage{english}


\subsubsection{Traditional Lexical Matching Models} In many of the previous competitions with legal case retrieval tasks, lexical matching models hold an important position due to their simplicity and interpretability in establishing the baseline for the model, without the extensive use of labeled training data. Hence, we implement the following lexical matching models:

\textbf{TF-IDF}:  TF-IDF\cite{juan2003tfidf} is a popular lexical matching model that employs word frequencies for the purpose of evaluating the importance of these words in relation to documents. It uses a combination of \textit{term frequency (TF)} and \textit{inverse document frequency (IDF)} as follows:
\begin{equation}
    TF(t_{i, j}) = \frac{n_{i, j}}{\sum_k n_{k, j}}
\end{equation}
\begin{equation}
    IDF(t_i) = \log\frac{|D|}{|D_i + 1|}
\end{equation}
\begin{equation}
    TF-IDF = TF \times IDF
\end{equation}
where D is the total number of documents in the corpus and $D_i$ represents the number of documents containing the word $t_i$. $n_{i,j}$ denotes the number of words $t_i$ in the document $d_j$.
TF-IDF has the disadvantage of the inability to carry semantic meaning in sentences. 

\textbf{BM25}: BM25\cite{stephen2009bm25} is a probabilistic model for ranking documents that relies on a bag-of-words approach. Given a query denoted as $q$ and a document represented as $d$, the BM25 formula is expressed in following:
\begin{equation}
    \operatorname{BM25}(d, q)=\sum_{i=1}^M \frac{\operatorname{IDF}\left(t_i\right) \cdot T F\left(t_i, d\right) \cdot\left(k_1+1\right)}{T F\left(t_i, d\right)+k_1 \cdot\left(1-b+b \cdot \frac{\operatorname{len}(d)}{\text { avgdl }}\right)}
\end{equation}

\textbf{QLD}: QLD \cite{chengxiang2008qld} stands as another probabilistic statistical model that can compute relevance scores using the probability of query generation. The QLD relevance score of a query $q$ and document $d$ is determined as follows:
\begin{equation}
\footnotesize
\log p(q \mid d)=\sum_{i: c\left(q_i ; d\right)>0} \log \frac{p_s\left(q_i \mid d\right)}{\alpha_d p\left(q_i \mid C\right)} 
+n \log \alpha_d+\sum_i \log p\left(q_i \mid C\right)
\end{equation}

\subsubsection{Fine-tuned BERT-based model}
With the advantages of contextual learning, BERT-based models are state-of-the-art in wide range of NLP tasks. Thus, a multingual BERT model is applied to produce sentence embeddings. Since, the multilingual BERT model is pre-trained on general domain data, we then fine-tune the model with ALQAC 2022 and 2023 datasets. 

\subsubsection{Classification models} 
After encoding a pair of question and candidate using a multingual BERT model, we feed these embeddings into other models including SVM, tree-based models.
This approach allows us to leverage the power of BERT for word embedding and then apply different classifiers to make specific predictions based on the vector representations. 

\subsubsection{Ensemble}
Finally, learning-to-rank techniques are employed to further enhance performance. In our method, we used LightGBM as ensemble model to integrate all the features into final score. The details of features are shown in Table \ref{tab:ensemble_features}.
LightGBM acts as an ensemble model that combines the outputs from different classifiers, taking into account their respective strengths and weaknesses, to produce a more robust and accurate final score. 
This approach ensures a comprehensive and well-informed decision based on the combined outputs from the different models used in the process. 

\begin{table}[ht]
\caption{Description of features used for learning-to-rank.}
\label{tab:ensemble_features}
\begin{tabular}{|c|l|p{5.75cm}|}
\hline
\textbf{Id} & \multicolumn{1}{c|}{\textbf{Name}} & \multicolumn{1}{c|}{\textbf{Description}}                                  \\ \hline
1           & BM25                               & Question-candidate scores with BM25 ranking                                \\ \hline
2           & TF-IDF                             & Question-candidate scores with TF-IDF                                      \\ \hline
3           & QLD                                & Question-candidate scores with QLD                                         \\ \hline
4          & BERT-cosine                        & Similarity scores between question and candidate vectors generated by BERT \\ \hline
5           & BERT-xgb                           & XGBoost scores of BERT embeddings                                          \\ \hline
6           & BERT-lgbm                          & LightGBM scores of BERT embeddings                                         \\ \hline
\end{tabular}
\end{table}




\subsection{Legal Question Answering} 
Given a legal question and relevant legal articles, the Legal Question Answering task aims to give the exact correct answer. This year, there are three types of questions in the competition including true/false, multiple choice, and span-based questions. We apply two approaches for this task including answer extraction for span-based questions, sentence classification for other questions.

\subsubsection{Span-based question}
The popular solution for span-based question answering problem is answer extraction. Given a question $q$ of $n$ tokens $q_1,q_2,...,q_n$ and a legal article $p$ of $m$ tokens $p_1,p_2,...,p_m$. The system aims to compute the probability $P(a|q,p)$ that each possible span $a$ is the answer. 

In the competition, we construct a question answering system based on \cite{nguyen2022vlh} work. The system consists of two main parts including sentence selection model and answer extraction model. The authors in \cite{nguyen2022vlh} show that their system is more effective than single model. We further fine-tune BERT-based models with ALQAC 2023 training dataset.

\subsubsection{True/False and Multiple choice questions}
In true/false questions, the input to the model is a legal passage and a question, the model aims to classify the question into a binary label. For multiple choice questions, the input to the model is a legal passage, a question and answer option. A multiple choice model aims to choose the correct answer among all the options based on the legal passage and question.

\begin{figure}[ht]
\centering
\includegraphics[width=\linewidth]{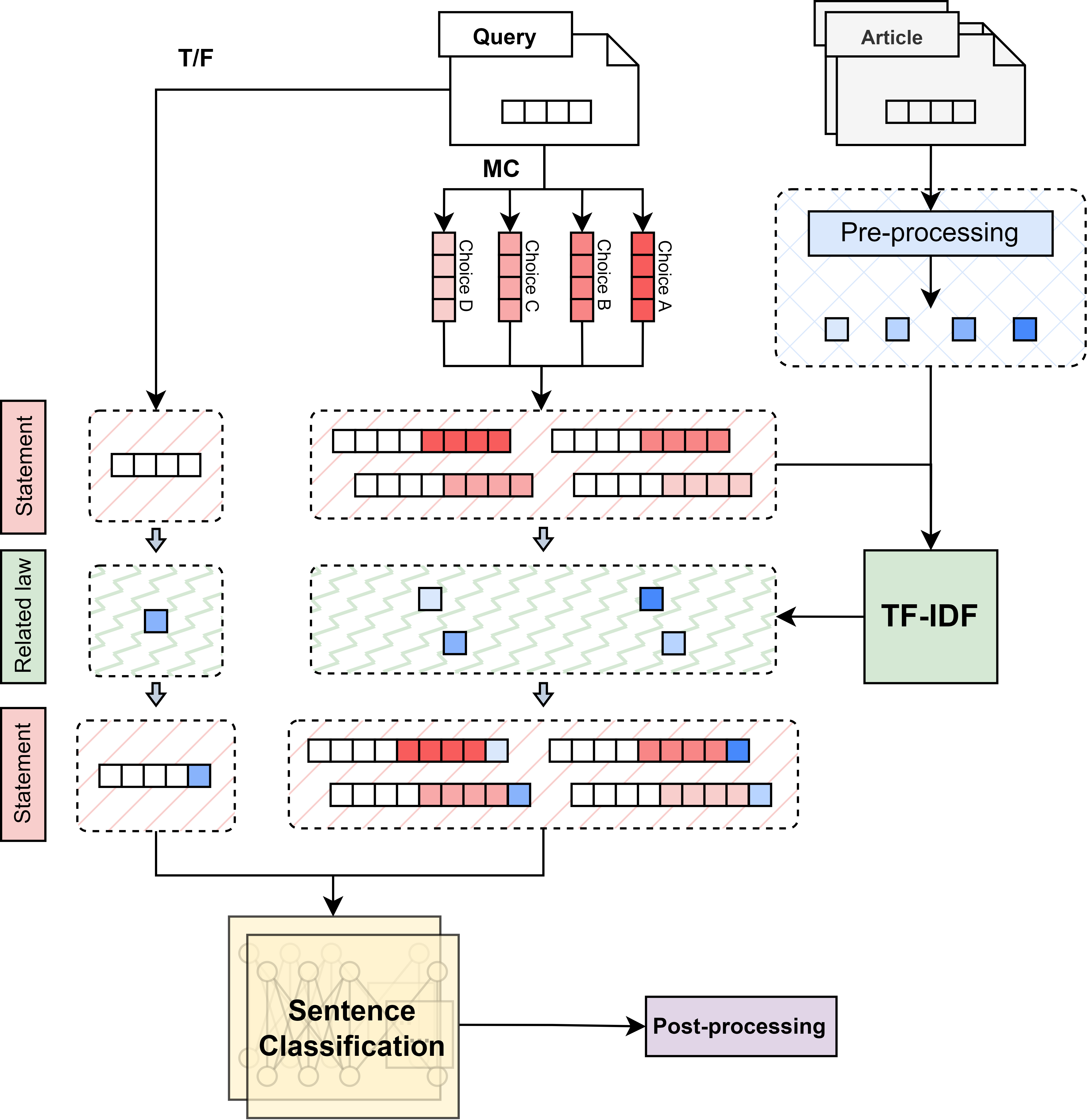}
\caption{True/false and multiple choice question answering system} 
\label{fig:task2_system}
\end{figure}

We consider these problems as a sentence classfication task. 
The details of our method for this task are presented in Table \ref{fig:task2_system}.
There are four main parts in our method as in the following order:
\begin{itemize}
    \item Data pre-processing.
    \item Question-clause matching.
    \item Sentence classfication. 
    \item Post-processing. 
\end{itemize}

\textbf{Data pre-processing:}
We find that the structure of a legal article is organized as follow: title of the article, clauses (begin with numbers) and sub-clauses (begin with letters). Therefore, legal articles are segmented into clause unit to overcome the input limitation of BERT-based models. 
The dataset in this competition consists of three types of questions: true/false, multiple choice and span-based questions. 
For multiple choice questions, we concatenate each option with the question to form a sequence. There are two types of multiple choice questions including: single answer and multiple answer. For multiple answer, we only concatenate options that contain information with the question. 

\textbf{Question-clause matching:}
TF-IDF is a popular algorithm that uses the frequency of words to determine how relevant those words are to a given document. This algorithm is applied to find the most related clause ans sub-clause for each question. Afterward, we concatenate question and retrieved clause and sub-clause into a statement.

\textbf{Sentence classification:}
BERT is a powerful deep learning model that has demonstrated state-of-the-art performance on a range of NLP tasks, including text classification. To enhance the mode comprehension of legal domain, we further fine-tune a multingual BERT model on ALQAC 2021 and 2023 datasets. 
As the differences in input format, we build two separate multilingual BERT models for answering true/false and multiple-choice questions. For true/false questions, an input is a pair of question and legal clause. For multiple choice questions, an input is a combination of question, option, and legal clause. 

\textbf{Post-processing:}
In case of single answer for multiple choice questions, the option with highest probability of 1 as the answer. Otherwise, if there are options ``both A and B is correct/wrong'' or ``all/none aboved options are correct'', we set a threshold to decide which answer to choose. 

\section{Experiments and Results}
\subsection{Data analysis}
The training dataset utilized for this year's tasks comprises official samples, encompassing 100 questions and 2131 articles distributed in 17 types of law. The organizer provides an additional Zalo legal dataset for document retrieval tasks, containing 3196 questions and 61425 articles. Table \ref{tab:length distribution} illustrates the text length distribution of each dataset. There is a wide variation in the article length. The longest article has more than 7500 words while the minimum number of words in an article is 17. The average length of an article is 854 words which exceeds the input limitation of BERT-based models. 
For the question-answering dataset, the average length of a query is approximately 110 words. However, there are some significantly long samples in the dataset. 

\begin{table}[ht]
\caption{Statistics of the ALQAC 2023 dataset}
\label{tab:length distribution}
\centering
\begin{tabular}{|l|cccc|}
\hline
\multicolumn{1}{|c|}{\multirow{2}{*}{\textbf{File}}} & \multicolumn{4}{c|}{\textbf{Words per sample}}                                            \\ \cline{2-5} 
\multicolumn{1}{|c|}{}                               & \multicolumn{1}{c|}{Min} & \multicolumn{1}{c|}{Max}  & \multicolumn{1}{c|}{Avg}    & Q3   \\ \hline
Legal corpus                                         & \multicolumn{1}{c|}{17}  & \multicolumn{1}{c|}{7535} & \multicolumn{1}{c|}{854.10} & 1066 \\ \hline
Question (training set)                              & \multicolumn{1}{c|}{25}  & \multicolumn{1}{c|}{286}  & \multicolumn{1}{c|}{111.38} & 144  \\ \hline
Question (test set)                                  & \multicolumn{1}{c|}{13}  & \multicolumn{1}{c|}{424}  & \multicolumn{1}{c|}{116.42} & 147  \\ \hline
\end{tabular}
\end{table}

\subsection{Legal Document Retrieval}

\indent The official evaluation metrics in the ALQAC 2023 for this task are listed below: 
\begin{equation*}
\begin{gathered}
\text {Precision}_i=\frac{\text {No. correctly retrieved articles of question } \mathrm{i}^{\mathrm{th}}} {\text {No. retrieved articles of question } \mathrm{i}^{\mathrm {th}}} \\
\text {Recall}_i=\frac{\text {No. correctly retrieved articles of question } \mathrm{i}^{\mathrm{th}}} {\text {No. relevant articles of question } \mathrm{i}^{\mathrm {th}}} \\
\mathrm{F2}_{\mathrm{i}}=\frac{5 \times \text {Precision}_{\mathrm{i}} \times \text { Recall}_{\mathrm{i}} } {4 \times \text { Precision}_{\mathrm{i}}+\text { Recall}_{\mathrm{i}}}\\
\mathrm{F2} =\text { average of }(\mathrm{F2}_{\mathrm{i}}) \\
\end{gathered}
\end{equation*} 

Due to the fact that we're working with Vietnamese data, we first apply RDRSegmenter algorithm to the queries and articles using VNCoreNLP Library\cite{vu-etal-2018-vncorenlp}. Additionally, because of the law data's characteristics (long and take a lot of time to search for related sentences), our team also divides the articles into smaller, easier-to-handle sections.

Fine-tuning our BERT model is also one of the main aspects in achieving high result to this task. The original BERT multilingual base model\footnote{https://huggingface.co/bert-base-multilingual-cased} was pre-trained on the 104 dialects with the largest Wikipedias, and is fundamentally aimed at being fine-tuned on tasks that utilize the entire sentence (potentially masked) to make decisions, such as sequence classification, token classification or question answering. Since the computational time of deep learning models is much slower than the BM25 ranking function, we decide to use BM25 due to its simplicity and decent results to generate negative samples. In the public test, we choose the first 50 articles retrieved by BM25 and take the irrelevant ones as negative samples. The training phase of BERT using contrastive learning is performed on GPU Tesla P100.

After that, we fed our data in the previously enhanced model to gain new $[CLS]$ embedding vectors and put them into the classification models. Below are the best possible scores achieved from the classification models after adjusting...:


Learning-to-rank techinque is employed to integrate all the scores from these models and the traditional lexical matching models. We achieve the first rank in the public test with F2 score of 0.94, as shown in Table \ref{tab:public_test_t1}. For public test, only top-1 relevant article is retrieved for each question.

\begin{table}[h]
\centering
\caption{Public Test Results for Task 1}
\label{tab:public_test_t1}
\begin{tabular}{|c|c|c|c|}
\hline
Team & Precision & Recall & F2-score \\
\hline
\textbf{NOWJ1} & \textbf{0.9400} & \textbf{0.9400} & \textbf{0.9400} \\
NeCo & 0.9200 & 0.9200 & 0.9200 \\
Sonic & 0.8900 & 0.8900 & 0.8900 \\
AI\_EPU & 0.8800 & 0.8800 & 0.8800 \\
Host\_ALQAC & 0.8700 & 0.8700 & 0.8700 \\
TechSquad & 0.3836 & 0.6900 & 0.5679 \\
\hline
\end{tabular}
\end{table}

For private test, we experiment 2 options including retrieve only top-1 article (run 1) and retrieve top-k articles (run  2, 3). Table \ref{tab:private_test_1} shows that the run 3 achieve highest F2 score among our 3 runs, and rank second place in the leaderboard. Further analysis reveals that the distribution of public and private datasets are a little different. The top-1 article method only achieve F2 score of 0.8298 in the private test. 

\begin{table}[ht]
\caption{Private test Results for Task 1}
\label{tab:private_test_1}
\begin{center}
\begin{tabular}{|c|c|c|c|}
\hline
Team & Precision & Recall & F2\\
\hline
NeCo & 0.9000 & 0.8621 & 0.8661 \\
\textbf{NOWJ1 \textit{(run 3)} } & \textbf{0.8636}  & \textbf{0.8348} & \textbf{0.8358}\\
\textbf{NOWJ1 \textit{(run 1)} } & \textbf{0.8636}  & \textbf{0.8258} & \textbf{0.8298}\\
\textbf{NOWJ1 \textit{(run 2)} } & \textbf{0.7955}  & \textbf{0.8394} & \textbf{0.8181}\\
Sonic \textit{(run 2)} & 0.8364 & 0.8136 & 0.8162 \\
AI\_EPU {\textit{(run 3)}} & 0.7091	& 0.6818 & 0.6848\\
ST \textit{(run 3)} & 0.2485 & 0.7227 & 0.5207 \\
\hline
\end{tabular}
\end{center}
\end{table}

\subsection{Legal Question Answering}

The main evaluation for this task is accuracy, which can be calculated with the formula below:
\begin{equation}
\begin{gathered}
\text {Accuracy}=\frac{\text {the number of correct answers}} {\text {the number of questions} }
\end{gathered}
\end{equation} 

ALQAC 2023 and 2021 datasets are used as training set, we validate our models on the ALQAC 2023 public test. Table \ref{tab:task2_public_test} shows our performance on the public test that contains only true/false and multiple choice questions. Our team achieve second place on the public leaderboard with Accuracy sore of 0.67.

\begin{table}[h]
\caption{Public test results for Task 2}
\label{tab:task2_public_test}
\begin{center}
\begin{tabular}{|c|c|c|c|}
\hline
Team & Accuracy\\
\hline
NeCo & 0.7400 \\
\textbf{NOWJ1} & \textbf{0.6700} \\
Sonic & 0.2900 \\
\hline
\end{tabular}
\end{center}
\end{table}

Table \ref{tab:task2_private_test} presents the private test leaderboard in the competition. The AIEPU team rank first place with Accuracy score of 0.8637. Our results show that the performance of multilingual BERT on legal question answering task is not impressive. This might due to the limitations of the dataset. Further research is needed to better understand the  limitations of our current model and to identify opportunities for improvement.

\begin{table}[h]
\caption{Private test Results for Task 2}
\label{tab:task2_private_test}
\centering
\begin{tabular}{|c|l|c|}
\hline
\textbf{Team} & \textbf{Submission(ID/Name)} & \textbf{Accuracy} \\
\hline
AIEPU & AIEPU\_submit\_top1.json & 0.8637 \\
Sonic & Sonic\_task2\_private\_test\_submission\_2 & 0.8000 \\
NeCo & NeCo\_run\_2.json & 0.7000 \\
\textbf{NOWJ1} & \textbf{NOWJ1\_run1\_sent\_classification.json} & \textbf{0.6545} \\
NOWJ1 & NOWJ1\_run\_3\_ens.json & 0.6455 \\
NOWJ1 & NOWJ1\_run2\_cos\_sim.json & 0.5818 \\
\hline
\end{tabular}
\end{table}

\section{Conclusions}
In this paper, we present our approaches in the ALQAC 2023 competition, which focuses on enhancing legal task performance with classic statistical models and pre-trained Language Models. In the competition, we achieved second place in the retrieval task and potential results in the question-answering task. 
Our experiments show that ensemble architectures have a significant performance in the retrieval task. Our experience has provided valuable insights into challenges and opportunities in the legal text processing domain. Further work could focus on exploring ensemble architecture and data enrichment to enhance the performance of our systems.



\bibliographystyle{IEEEtranN}
\bibliography{ref}

\end{document}